\documentclass[11pt]{article}

\usepackage[preprint]{acl}
\usepackage{times}
\usepackage{latexsym}
\usepackage[T1]{fontenc}
\usepackage[utf8]{inputenc}
\usepackage{microtype}
\usepackage{inconsolata}
\usepackage{graphicx}
\usepackage{amsmath}
\usepackage{amssymb}
\usepackage{booktabs}
\usepackage{float}

\title{GVD: Governed Versioning and Deduplication\\
for Document Repositories}

\author{
  Mohammadreza Sediqin \\
  Centific Research \\
  New York, USA \\
  \texttt{mohammadreza.s@centific.com} \\\And
  Shivali Dalmia \\
  Centific Research \\
  Seattle, USA \\
  \texttt{shivali.dalmia@centific.com} \\\AND
  Sumukha Thoppanahalli \\
  Centific Research \\
  Seattle, USA \\
  \texttt{sumukhasharma.t@centific.com} \\\And
  Abhishek Mukherji \\
  Centific Research \\
  Seattle, USA \\
  \texttt{abhishek.mukherji@centific.com}
}

\begin{document}

\maketitle
\begin{abstract}
Document repositories evolve continuously. Guidelines and policies are revised,
superseded, and re-uploaded, so the same content recurs in different wording and
newer versions refine or contradict earlier ones. These inconsistencies belong to
the growing collection rather than to any single document, yet existing work
treats versioning, duplicate detection, and contradiction detection as isolated
pairwise tasks and stops once a pair is labeled. We present GVD (Governed Versioning and Deduplication), a framework that unifies
cross-document version linking with rule-level conflict resolution under an
auditable update policy. Incoming
documents are assigned to version families through bidirectional rule alignment,
and their rules are compared against the family memory to identify duplicates,
contradictions, asymmetric refinements, and new knowledge, with Counterfactual
Span Probing (CSP) resolving related pairs that inference misclassifies as
neutral. Relation-specific policies suppress duplicates and escalate only
consequential changes for review, retaining version lineage as an audit trail.
The pipeline runs fully locally, with no large language model. On 120 enterprise
documents processed as 140 ingestions across 59 version families, GVD reaches an
F1 of 0.97 for version-family construction and 0.94 for rule-level consistency,
with CSP raising rule consistency from 0.90 to 0.94.
\end{abstract}

\section{Introduction}
Document repositories do not stand still. In enterprise, legal, financial, healthcare, and government settings, guidelines, operational policies, and procedural specifications are revised, superseded, and re-uploaded as requirements change, so a repository accumulates several versions of the same document alongside overlapping and sometimes conflicting guidance. Repositories also contain multimodal documents with text, tables, figures, charts, and images \cite{gao2026scaling,dalvand2025indian, dalvand2025regional}. Duplicates add redundancy, contradictions add ambiguity, and unmanaged revisions make it hard to tell which version of a requirement to trust. In regulated settings this is not only a quality problem. An organization needs to know which version of a rule was in force, what replaced it, and who approved the change.

Existing work treats document similarity, duplicate detection, and contradiction detection as separate pairwise tasks over fixed collections, and stops once a pair has been labeled. Governing a live repository asks more. Incoming content must be linked to the history it revises, reconciled with what is already held, and recorded in a way that can be inspected later. Some of those decisions are routine and some are consequential enough to need a person, and telling the two apart is what keeps review effort affordable as the collection grows.

We present GVD (Governed Versioning and Deduplication), which represents document content as normalized rules and governs the collection in two stages. Stage~1 assigns each incoming document to a version family through bidirectional rule alignment, so a short revision is not absorbed into a larger family. Stage~2 compares its rules against the family memory and labels each pair as duplicate, contradiction, asymmetric refinement, or new knowledge. Relation-specific policies then act on those labels: duplicates are suppressed, new knowledge is inserted, and only consequential changes are escalated for review, with every accepted change linked to the revision that introduced it. The pipeline runs locally on encoder-scale models, with no large language model, so document content never leaves the deployment environment.

To our knowledge, GVD is the first framework to turn consistency relations into repository update policy over an evolving collection, rather than stopping at classification. Our contributions are:

\begin{enumerate}
\item A version-linking method that groups documents into families by bidirectional rule coverage, robust to revisions that differ in length and wording.
\item A governed update policy that acts on consistency relations, escalating only consequential changes for review and retaining version lineage as an audit trail, with routing decisions validated against expert review.
\item Counterfactual Span Probing (CSP), a lightweight training-free mechanism that separates cosmetic rewording from substantive edits, keeping reworded duplicates from entering the collection as new rules.
\end{enumerate}

\section{Related Work}

\textbf{Document similarity and versioning.}
Lexical fingerprinting methods such as shingling, SimHash~\citep{williams2013near},
and Sectional MinHash~\citep{hassanian2018sectional} degrade under paraphrasing
and produce binary duplicate decisions without modeling version families.
Semantic approaches instead apply word-embedding distances to unsupervised
revision detection~\citep{zhou2019text,zhu2017semantic,11248014} and contextual embeddings
to sentence-level alignment across versions~\citep{spangher2024llms,sediqin2025laces}, but operate
at coarse granularity and do not enforce bidirectional consistency. GVD operates
at the rule level and requires bidirectional agreement.

\textbf{Deduplication and contradiction detection.}
Because pure similarity cannot distinguish semantic closeness from logical
conflict~\citep{das2021sentence,shambour2022effective}, work in requirements
engineering turns to natural language inference (NLI) to classify pairs as
duplicate, conflict, or neutral~\citep{malik2023transfer}, with hybrid embedding
models~\citep{saleem2026reqnet} and retrieve-and-classify pipelines for duplicate
bug reports~\citep{ariai2025natural}. These treat detection as pairwise
classification and do not model repository-level decisions such as whether a rule
should be retained, suppressed, escalated, or added as new. GVD extends semantic
classification into a repository-maintenance workflow combining retrieval,
bidirectional inference, asymmetric relation detection, and update policies.

\textbf{Human-in-the-loop review.}
Automatic classification leaves ambiguous cases that require human
judgment~\citep{de2020lie}, yet prior work handles document-level duplication and
rule-level inconsistency in isolation, without governing an evolving
collection~\citep{surana2022identifying, dalmia2026guide}, and human-in-the-loop
pipelines~\citep{wu2022survey} stop after classification rather than acting on it.
GVD resolves ambiguous cases automatically where possible and routes
contradictions, asymmetric refinements, and value or polarity changes for
validation, handling routine updates without intervention.

\section{System Framework}

GVD is a governance framework for evolving document repositories. As documents are revised, re-uploaded, and extended, the repository accumulates duplicates, conflicting instructions, overlapping guidance, and multiple versions of related processes. Where existing approaches treat versioning, duplicate detection, and contradiction analysis as independent tasks, GVD unifies them into a single update policy, evaluating incoming content before it is incorporated so the collection stays consistent as it evolves.

\begin{figure*}[t]
\centering
\includegraphics[width=0.8\textwidth]{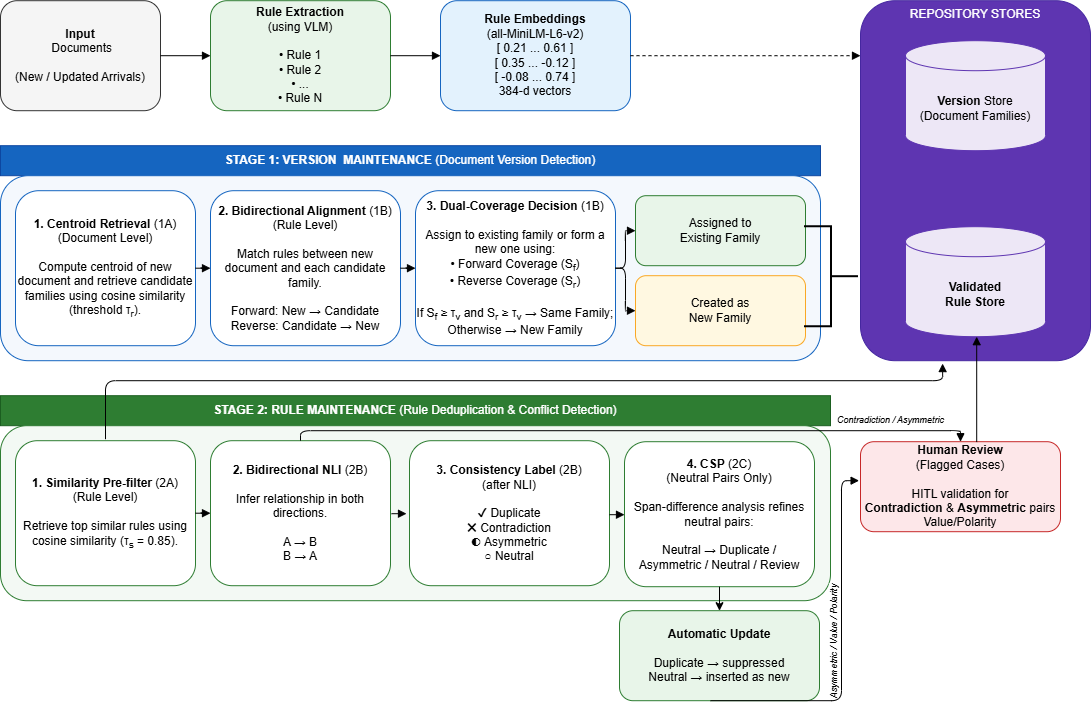}
\caption{Overview of GVD. Incoming documents are converted into validated
RuleUnits and processed in two stages: (1) version maintenance assigns documents
to families, and (2) rule maintenance classifies rules into consistency relations
via bidirectional NLI and CSP. Relation-specific policies then update the store,
suppressing duplicates automatically and routing consequential changes for human
review.}
\label{fig:arch}
\end{figure*}

Rather than operating on raw documents, GVD reasons over a shared rule-level representation, which enables finer-grained consistency reasoning while reducing sensitivity to formatting and paraphrasing. Incoming PDF, DOCX, and PPTX documents are first converted into \texttt{RuleUnit}s by an independent multimodal preprocessing pipeline and approved through human-in-the-loop review prior to ingestion; GVD therefore operates only on validated, human-approved rules and focuses on maintaining consistency within the store rather than generating or validating the rules themselves. Formally, a document is $D=\{\mathbf{r}_1,\mathbf{r}_2,\ldots,\mathbf{r}_N\}$
where each rule embedding $\mathbf{r}_i \in \mathbb{R}^{384}$ is generated using the all-MiniLM-L6-v2 sentence encoder, which provides a lightweight 384-dimensional representation while maintaining strong semantic retrieval performance. The same encoder is used throughout both stages to ensure a consistent embedding space for document retrieval, version matching, and rule-level consistency analysis, as illustrated in Figure~\ref{fig:arch}.

\subsection{Stage 1: Version Maintenance}

The objective of this stage is to determine whether an incoming document belongs to an existing version family or represents new knowledge that should initialize a new family. This is non-trivial because successive versions of the same guideline document often differ substantially in content and rule count: a new revision may add, remove, or rephrase rules, so two documents in the same family need not contain the same number of rules or identical text. Stage 1 therefore proceeds in two steps, a fast centroid-based retrieval followed by a stricter bidirectional alignment.

\textbf{Stage 1A: Centroid retrieval.} Each document is summarized by the centroid of its rule embeddings,
\begin{equation}
\mathbf{c}_D=
\frac{1}{|D|}
\sum_{i=1}^{|D|}
\mathbf{r}_i ,
\label{eq:centroid}
\end{equation}
where $|D|$ is the number of rules in the document. For an incoming document $D_q$ and a stored document $D_j$, centroid similarity is computed by cosine similarity,
\begin{equation}
\mathrm{sim}(D_q,D_j)=
\frac{\mathbf{c}_{D_q}\cdot\mathbf{c}_{D_j}}
{\lVert\mathbf{c}_{D_q}\rVert\,\lVert\mathbf{c}_{D_j}\rVert}.
\label{eq:docsim}
\end{equation}
The top-$k$ documents exceeding a retrieval threshold are retained as candidates, shrinking the search space before the more expensive rule-level comparison.

\textbf{Stage 1B: Bidirectional alignment.} Each candidate then undergoes rule-level alignment, where the similarity between two rules is
\begin{equation}
s(\mathbf{r}_i,\mathbf{r}_j)=
\frac{\mathbf{r}_i\cdot\mathbf{r}_j}
{\lVert\mathbf{r}_i\rVert\,\lVert\mathbf{r}_j\rVert}.
\label{eq:rulesim}
\end{equation}
Forward coverage measures how many rules in the incoming document are represented in the candidate,
\begin{equation}
S_f=
\frac{1}{|D_q|}
\sum_{\mathbf{r}_i \in D_q}
\mathbb{I}
\!\left(
\max_{\mathbf{r}_j \in D_c}
s(\mathbf{r}_i,\mathbf{r}_j)\ge\tau
\right),
\label{eq:forward}
\end{equation}
while reverse coverage measures the converse,
\begin{equation}
S_r=
\frac{1}{|D_c|}
\sum_{\mathbf{r}_j \in D_c}
\mathbb{I}
\!\left(
\max_{\mathbf{r}_i \in D_q}
s(\mathbf{r}_i,\mathbf{r}_j)\ge\tau
\right).
\label{eq:reverse}
\end{equation}
The incoming document joins an existing family through a dual-coverage decision: it joins only when both directional coverage scores exceed the version threshold,
\begin{equation}
(D_q,D_c)\in F
\iff
S_f\ge\tau_v
\;\land\;
S_r\ge\tau_v .
\label{eq:family}
\end{equation}
Requiring agreement in both directions, rather than one-directional similarity, prevents a small set of rules from being wrongly absorbed into a larger family and gives a more reliable picture of document evolution even when versions differ in length or wording.

\subsection{Stage 2: Rule Maintenance}

After version assignment, the incoming document enters its version family and its rules are compared against the family memory, which consists of all validated rules previously associated with that lineage, including rules from earlier revisions, historical versions, and accepted repository updates. Consistency decisions are therefore made not only against the current document but against the accumulated knowledge of the entire version family. Stage 2 proceeds in three steps: a cosine similarity pre-filter that retrieves candidate rules, bidirectional inference that assigns a consistency label, and Counterfactual Span Probing that resolves the residual neutral pairs.

\textbf{Stage 2A: Similarity pre-filter.} For each incoming rule, cosine similarity identifies the most relevant candidate rule in the family memory. A rule whose best cosine similarity falls below a threshold $\tau_s$ has no close match in the store and is inserted directly as new knowledge; only pairs at or above $\tau_s$ proceed to inference. We set $\tau_s = 0.85$.

\textbf{Stage 2B: Bidirectional inference.} For each candidate pair, inference is performed with DeBERTa-v3-large fine-tuned on SNLI and MultiNLI, applied in both directions. Let $P_e^{\rightarrow}$ and $P_e^{\leftarrow}$ denote the entailment probabilities of the existing-to-new and new-to-existing directions, and $P_c$ the contradiction probability. The pair is labeled as
\begin{equation}
\mathrm{Label}=
{\small
\begin{cases}
\text{Contradiction}, & \max(P_c^{\rightarrow},P_c^{\leftarrow})\ge\tau_c,\\[2pt]
\text{Duplicate}, & P_e^{\rightarrow}\ge\tau_d \land P_e^{\leftarrow}\ge\tau_d,\\[2pt]
\text{Asymmetric}, & (P_e^{\rightarrow}\ge\tau_d)\oplus(P_e^{\leftarrow}\ge\tau_d),\\[2pt]
\text{Neutral}, & P_e^{\rightarrow}<\tau_d \land P_e^{\leftarrow}<\tau_d,
\end{cases}}
\label{eq:nli}
\end{equation}
where $\rightarrow$ and $\leftarrow$ denote the directions $r_{\text{exist}}\!\rightarrow\!r_{\text{new}}$ and $r_{\text{new}}\!\rightarrow\!r_{\text{exist}}$, and $\oplus$ denotes exclusive or. A pair is a Duplicate when entailment holds in both directions, Asymmetric when it holds in exactly one direction, capturing the refinements and partial extensions that frequently occur as documents evolve, and Neutral when neither direction entails and there is no contradiction.

\textbf{Stage 2C: Counterfactual Span Probing.} Bidirectional inference reliably identifies clear duplicates and contradictions, but many semantically related rules fall into the Neutral class, where hidden duplicates and asymmetric refinements remain unresolved. CSP is a training-free refinement applied only to neutral pairs, designed to recover latent duplicate and asymmetric relations that single-pass inference collapses into the neutral class. Rather than re-scoring entire rules, CSP isolates the spans responsible for disagreement and probes, counterfactually, whether those differences are substantive or merely cosmetic. This allows GVD to distinguish cosmetic reformulations from substantive modifications without additional model training. The identified spans are categorized as cosmetic rewordings, value or numerical changes, polarity reversals, or one-sided clauses, which are used to refine the final consistency label. Let $V$, $P$, $C_1$, and $C_2$ denote, respectively, the presence of any value or numerical change, any polarity reversal, a substantive one-sided clause on exactly one side, and substantive clauses on both sides. The pair is relabeled as
\begin{equation}
\mathrm{CSP}=
\begin{cases}
\text{Review}, & V \lor P,\\[2pt]
\text{Neutral}, & \lnot(V \lor P) \land C_2,\\[2pt]
\text{Asymmetric}, & \lnot(V \lor P \lor C_2) \land C_1,\\[2pt]
\text{Duplicate}, & \text{otherwise}.
\end{cases}
\label{eq:csp}
\end{equation}
A pair is labeled Asymmetric only when a substantive one-sided clause is its sole non-cosmetic difference; a value or polarity change is instead routed to human review as a substantive edit to an existing rule, while a two-sided change with no value or polarity edit remains Neutral. A one-sided clause is substantive when, after discarding short function-like tokens (fewer than three characters, such as ``if'', ``to'', or ``a''), at least three content tokens remain. Both thresholds were set on the validation subset: lower values admit trivial additions as spurious refinements, while higher values discard genuine clause-level changes. The inference model is re-applied only on replacement spans, as a targeted polarity and contradiction check, rather than as the labeling rule; the Duplicate and Asymmetric decisions themselves rest on span structure and token counts and require no further inference call. CSP operates only on residual neutral pairs, routing value and polarity edits to review and otherwise emitting Duplicate, Asymmetric, or Neutral. Contradictions are detected upstream by Eq.~\ref{eq:nli}, which classifies them reliably (F1~0.97).
\begin{table*}[t]
\centering
\small
\setlength{\tabcolsep}{6pt}
\renewcommand{\arraystretch}{1.05}
\begin{tabular}{llcc}
\toprule
 & Rule pair & NLI & CSP \\
\midrule
Ex.\ 1 & A: ``commercial is announced and heard'' & Neutral & Duplicate \\
       & B: ``commercial is announced and follows'' & & \\
\addlinespace
Ex.\ 2 & A: ``estimate and report speaker count'' & Neutral & Asymmetric \\
       & B: ``estimate and report speaker count using best judgment'' & & \\
\bottomrule
\end{tabular}
\caption{Two CSP reclassification examples.}
\label{tab:csp_examples}
\end{table*}
Table~\ref{tab:csp_examples} shows two such cases: a cosmetic rewording relabeled Duplicate, and a substantive added clause relabeled Asymmetric.

\subsection{Repository Update and Human Validation}

Following consistency classification, repository updates follow relation-specific policies. Duplicate rules are suppressed, and new rules are inserted directly. Contradictory rules, asymmetric rules, and rules carrying a value or polarity change are quarantined and routed for human review before repository modification, while asymmetric and reviewed edits are stored as refinements linked to their parent rule, preserving version lineage while preventing conflicting values from silently entering the store. By combining rule-level representation, version-family maintenance, bidirectional consistency reasoning, CSP, and selective human validation within a single framework, GVD governs how a document repository changes over time while keeping every accepted update traceable to the revision that introduced it.

\section{Experiments}

\subsection{Dataset and Ground Truth}
Because evolving enterprise repositories with expert validation are hard to obtain at scale, we prioritize a real, expertly annotated corpus over a larger synthetic one. Our evaluation corpus consists of 120 enterprise guideline documents provided by
industrial clients under confidentiality agreements, in PDF, DOCX, and PPTX
format, processed into approximately 3,900 validated RuleUnits. The corpus was replayed as 140 repository ingestions to simulate evolution: 59 first uploads initializing version families, 61 revised versions of existing documents, and 20 exact reuploads, forming 59 version families and yielding 4,396 rule-level consistency decisions. Since enterprise guideline repositories provide no ground-truth labels, we construct balanced evaluation sets through expert human annotation. For Stage 1, we sample 300 document pairs across the 59 version families, balanced into 150 same-family and 150 different-family pairs, ensuring balanced evaluation of grouping and separation. For Stage 2, we annotate 500 rule pairs across four consistency relations (130 duplicate, 122 contradiction, 128 asymmetric, 120 neutral). Each set was independently labeled by expert annotators familiar with the guidelines, achieving 0.93 agreement with Cohen's $\kappa = 0.88$ (Stage 1) and 0.90 agreement with $\kappa = 0.85$ (Stage 2).

\subsection{Experimental Setup}
GVD uses the all-MiniLM-L6-v2 sentence encoder to produce rule embeddings for both document-level and rule-level retrieval, and DeBERTa-v3-large fine-tuned for Natural Language Inference for bidirectional semantic inference. All thresholds were tuned on a small held-out validation subset, disjoint from the Stage 1 and Stage 2 evaluation sets to avoid leakage, and then fixed for all reported experiments. For Stage 1A candidate retrieval, we evaluated $k \in \{3, 5, 8, 10\}$ on this validation subset and fixed $k = 5$; $k = 3$ reduced family coverage, while $k = 8$ and $k = 10$ added only redundant candidates that bidirectional alignment subsequently removed. For Stage~1, we set the centroid threshold $\tau_r = 0.70$, per-rule match $\tau = 0.75$, and version threshold $\tau_v = 0.75$; for Stage~2, $\tau_s = 0.85$, entailment $\tau_d = 0.70$, and contradiction $\tau_c = 0.85$. All experiments run on a single NVIDIA GPU. Stage~1 is a binary decision, reported as precision, recall, and F1. For Stage~2, the precision, recall, and F1 values reported in Table~\ref{tab:stage2_all} are macro-averaged across the four consistency relations (Duplicate, Contradiction, Asymmetric, and Neutral).

\subsection{Versioning and Consistency Results}
On the 300 labeled document pairs, GVD reaches an F1 of 0.971 for version grouping (Table~\ref{tab:stage1_all}), correctly placing same-family documents together and keeping unrelated ones apart. On the 500 labeled rule pairs, it reaches precision 0.97, recall 0.91, and F1 0.94 (Table~\ref{tab:stage2_all}). Bidirectional inference handles clear duplicates and contradictions well but leaves many related rules in the neutral class, where hidden duplicates and refinements would otherwise enter the store as new. CSP revisits these neutral pairs and raises rule-consistency F1 from 0.90 to 0.94, with the largest gains on the duplicate and neutral classes (Table~\ref{tab:csp_effect}).

\begin{table}[t]
\centering
\small
\setlength{\tabcolsep}{6pt}
\renewcommand{\arraystretch}{1.05}
\begin{tabular}{lccc}
\toprule
\textbf{Class} & \textbf{F1 (no CSP)} & \textbf{F1 (+CSP)} & \textbf{$\Delta$} \\
\midrule
Duplicate     & 0.85 & 0.93 & $+0.08$ \\
Contradiction & 0.97 & 0.97 & --- \\
Asymmetric    & 0.91 & 0.92 & $+0.01$ \\
Neutral       & 0.85 & 0.92 & $+0.07$ \\
\midrule
F1      & 0.90 & 0.94 & $+0.04$ \\
\bottomrule
\end{tabular}
\vspace{-6pt}
\caption{Rule-relation classification: effect of CSP, per class and overall.}
\label{tab:csp_effect}
\end{table}

\subsection{Baselines and Ablations}
Because no prior system directly targets this task, we evaluate standard baselines and ablations on our labeled datasets. For Stage 1, we compare lexical fingerprinting methods (MinHash and Shingling with Jaccard~\citep{broder1997resemblance}, and SimHash~\citep{charikar2002similarity}) and embedding-based similarity (SBERT cosine and clustering~\citep{reimers2019sentence}). We additionally compare against RETSim~\citep{zhang2024retsim}, a lightweight model purpose-built for near-duplicate and versioning detection, which embeds text with a character-level vectorizer trained via metric learning and compares embeddings by cosine similarity. For all score-based baselines, the same-family decision threshold was selected on
the same held-out validation subset used to tune GVD's thresholds, so that each
method is evaluated at its own best operating point. The lexical and general-embedding baselines reach only 0.62 to 0.70 F1 (SBERT cosine strongest at 0.704). RETSim is far stronger, reaching 0.943, reflecting its specialization for near-duplicate detection. It nonetheless still trails GVD (0.971): because RETSim scores document similarity directly, it does not model the bidirectional, family-level coverage that reliably separates same-family from different-family documents, particularly when versions differ in length or wording. The ablation then isolates the matching rule: centroid-only similarity is too coarse (0.837), one-directional coverage still over-merges short documents (0.944), and bidirectional coverage gives the best result (0.971) (Table~\ref{tab:stage1_all}).
\begin{table*}[t]
\centering
\small
\setlength{\tabcolsep}{6pt}
\renewcommand{\arraystretch}{1.05}
\begin{tabular}{@{}llc@{}}
\toprule
\textbf{System} & \textbf{Method} & \textbf{Reported F1} \\
\midrule
NewsEdits    & Embedding sentence alignment           & 0.95 \\
arXivEdits   & Neural CRF sentence alignment          & 0.94 \\
FSARC        & Semantic eight-tuple + heuristic rules & 0.91 \\
Malik et al. & Encoder with transfer learning         & 0.91 \\
PassionNet   & LLM with similarity                    & 0.93 \\
LegalWiz     & NLI with LLM judge                     & 0.71 \\
\bottomrule
\end{tabular}
\caption{Prior systems on their own datasets, not directly comparable to GVD.}
\label{tab:baselines}
\end{table*}
For Stage 2 (Table~\ref{tab:stage2_all}), a vanilla NLI baseline that maps a single forward pass directly to a label reaches 0.72. The component ablation isolates each part: cosine similarity alone reaches only 0.56, since it cannot separate logical relations from surface closeness; adding bidirectional NLI raises F1 to 0.90; and CSP lifts recall from 0.84 to 0.91 (F1 0.94) by recovering hidden duplicates and asymmetric refinements left in the neutral class. Lexical methods such as MinHash and SimHash apply only to Stage 1, as they yield similarity scores and cannot express contradiction or asymmetric relations; for Stage 2 we therefore compare against cosine-only and vanilla NLI baselines.

\begin{table}[t]
\centering
\small
\setlength{\tabcolsep}{4pt}
\renewcommand{\arraystretch}{1.05}
\begin{tabular}{lccc}
\toprule
Method & P & R & F1 \\
\midrule
\multicolumn{4}{l}{\textit{Standard baselines}} \\
MinHash          & 0.58 & 0.67 & 0.621 \\
Shingling (word) & 0.61 & 0.67 & 0.637 \\
SimHash          & 0.57 & 0.73 & 0.642 \\
Shingling (char) & 0.60 & 0.71 & 0.652 \\
SBERT clustering & 0.66 & 0.73 & 0.693 \\
SBERT cosine     & 0.68 & 0.73 & 0.704 \\
RETSim           & 0.917 & 0.970 & 0.943 \\
\midrule
\multicolumn{4}{l}{\textit{GVD coverage ablation}} \\
Centroid only        & 0.800 & 0.878 & 0.837 \\
One-directional      & 0.931 & 0.957 & 0.944 \\
Bidirectional (full) & 0.969 & 0.974 & \textbf{0.971} \\
\bottomrule
\end{tabular}
\caption{Stage 1 version grouping (300 pairs).}
\label{tab:stage1_all}
\end{table}

\begin{table}[t]
\centering
\small
\setlength{\tabcolsep}{4pt}
\renewcommand{\arraystretch}{1.05}
\begin{tabular}{lccc}
\toprule
Method & P & R & F1 \\
\midrule
\multicolumn{4}{l}{\textit{Standard baseline}} \\
Vanilla NLI         & 0.69 & 0.75 & 0.72 \\
\midrule
\multicolumn{4}{l}{\textit{GVD ablation}} \\
Cosine only         & 0.70 & 0.47 & 0.56 \\
Cosine + NLI        & 0.97 & 0.84 & 0.90 \\
Cosine + NLI + CSP  & 0.97 & 0.91 & \textbf{0.94} \\
\bottomrule
\end{tabular}
\caption{Stage 2 rule consistency (500 pairs).}
\label{tab:stage2_all}
\end{table}
\subsection{Relation Detection on the Evolving Store}

Across all 140 document ingestions, each incoming rule is either inserted directly as new knowledge when no sufficiently similar rule exists in the family memory, or compared against its nearest retrieved candidate and assigned a consistency relation. In total, GVD processed 4{,}396 rule-level decisions, yielding 2{,}719 duplicate relations, 1{,}173 new-rule insertions, 194 asymmetric refinements, 187 contradictions, and 123 neutral cases after CSP refinement.
Duplicate relations dominate, reflecting the substantial overlap that naturally occurs across document revisions and reuploads. Contradictions and asymmetric refinements, including those carrying a value or polarity change, are routed for human validation, giving 381 cases for review. The remaining neutral cases are genuinely new and inserted directly.
\subsection{Routing Policy Evaluation}
To evaluate the governance policy rather than relation classification alone, we
manually reviewed the cases routed for human validation, comprising
contradictions, asymmetric refinements, and value or polarity edits. Reviewers
confirmed that these cases required human judgment before repository
modification, and a sample of automatically handled decisions showed no instance
in which a contradiction, refinement, or value or polarity change bypassed
review. The governance layer therefore concentrates human effort on consequential
updates while automating routine ones.
\subsection{Comparison with Prior Approaches}
Table~\ref{tab:baselines} lists prior systems for versioning, deduplication, and contradiction detection. Because they are evaluated on different datasets and tasks, we do not compare their scores directly against GVD; we instead summarize each method and the result it reports on its own benchmark.
For document versioning and near-duplicate detection, NewsEdits~\citep{spangher2024newsedits} aligns sentences across document versions with contextual embeddings to detect revisions, improving cross-version sentence linking over its earlier version. arXivEdits~\citep{jiang2022arxivedits} trains a neural CRF to align sentences across multiple revisions of scientific papers, reaching high alignment F1 on its own revision corpus. For rule-level consistency, FSARC~\citep{guo2021automatically} parses requirements into a semantic eight-tuple and applies heuristic rules to flag conflicts, achieving high recall on natural-language requirements without supervised training. \citet{malik2023transfer} use sequential transfer learning over transformer encoders to classify requirement pairs as duplicate, conflict, or neutral, improving accuracy in data-rich settings. PassionNet~\citep{saleem2025passionnet} combines an LLM with similarity features to identify duplicate and conflicting requirements, while LegalWiz~\citep{mantravadi2025legalwiz} pairs NLI with an LLM judge to detect contradictions in legal text.

These systems share two limitations that GVD addresses: versioning and near-duplicate methods operate only at the document or sentence level and cannot express contradiction or asymmetric refinement, while consistency methods treat detection as one-off pairwise classification without modeling repository-level updates over an evolving store. GVD works at the rule level, unifies versioning and consistency within a single pipeline, and resolves ambiguous neutral pairs through training-free CSP, which recovers latent relations missed by single-pass inference without relying on a large language model.

\section{Conclusion}

We presented GVD, a fully local framework for governing how a document repository changes as it accumulates revisions. GVD groups documents into version families through centroid retrieval and bidirectional rule alignment, and manages rule-level consistency through inference and Counterfactual Span Probing, escalating only consequential changes for human review and resolving uncertain neutral cases without any large language model. On a real enterprise corpus of 120 documents, it reaches an F1 of 0.97 for versioning and 0.94 for rule consistency, with CSP raising F1 from 0.90 to 0.94. Our results show that governing an evolving collection is a distinct problem from pairwise duplicate or contradiction detection, and that acting on consistency relations rather than stopping at them keeps the repository auditable as it grows. In future work, we plan to extend GVD across enterprise domains and languages, evaluate it on larger and more diverse rule sets, and integrate it into a retrieval-augmented pipeline to measure its effect on downstream answer quality.



\section*{Limitations}
The present evaluation covers English enterprise guideline documents, chosen
because expert validation was available at the scale required for reliable ground
truth; extension to further languages and document domains is a natural next
step. GVD is scoped to the governance layer and operates on validated rules
supplied by an upstream extraction pipeline, which keeps versioning and
consistency decisions independent of any particular parser. It is evaluated on
the governance decisions themselves, through version-family construction,
relation classification, and routing validation, so its effect on downstream
retrieval and generation is a natural direction for future work.


\bibliography{reference}

\end{document}